\documentclass[10pt,twocolumn,letterpaper]{article}

\usepackage{iccv}
\usepackage{times}
\usepackage{epsfig}
\usepackage{graphicx}
\usepackage{amsmath}
\usepackage{amssymb}
\usepackage{bm}
\usepackage{booktabs} 
\usepackage{subcaption}
\usepackage{wrapfig}
\usepackage{multirow}
\usepackage[abs]{overpic}

\usepackage[dvipsnames]{xcolor}

\newcommand{\para}[1]{\vspace{0.5mm}\noindent\textbf{{#1}.}~~}

\usepackage[pagebackref=true,breaklinks=true,letterpaper=true,colorlinks,bookmarks=false]{hyperref}

\iccvfinalcopy 

\def\iccvPaperID{4835} 

\begin{document}

\title{Floorplan-Jigsaw: Jointly Estimating Scene Layout and Aligning Partial Scans}


\author{
\vspace{2mm}Cheng Lin \quad\quad Changjian Li \quad\quad Wenping Wang\\
\text{The University of Hong Kong}\\
{\tt\small \{clin, cjli, wenping\}@cs.hku.hk}
}

\maketitle

\begin{abstract}
We present a novel approach to align partial 3D reconstructions which may not have substantial overlap. Using floorplan priors, our method jointly predicts a room layout and estimates the transformations from a set of partial 3D data. Unlike the existing methods relying on feature descriptors to establish correspondences, we exploit the 3D ``box'' structure of a typical room layout that meets the Manhattan World property. We first estimate a local layout for each partial scan separately and then combine these local layouts to form a globally aligned layout with loop closure. Without the requirement of feature matching, the proposed method enables some novel applications ranging from large or featureless scene reconstruction and modeling from sparse input. We validate our method quantitatively and qualitatively on real and synthetic scenes of various sizes and complexities. The evaluations and comparisons show superior effectiveness and accuracy of our method. 
\end{abstract}




\section{Introduction}
\label{sec:intro}
Indoor scene understanding and reconstruction have been extensively researched in computer vision. In recent years, the development of consumer RGB-D sensors has greatly facilitated 3D data capture and enabled high-quality reconstruction of indoor scenes. Although many methods have been proposed for continuous camera localization to register 3D depth data, it remains a challenge to scan some scenes in a single pass. The main difficulty is caused by interruptions in camera tracking, which results in a number of partial scans with little overlap. This frequently occurs in the following typical scenarios: (1) a large-scale scene is scanned region-by-region rather than in a single pass to reduce the workload or to meet the memory limit of a computer; (2) when scanning featureless areas or doorways, camera tracking often fails and so leads to several partial scans without sufficient overlap or feature points; (3) when a large scene is scanned using multiple robots, the scene is usually explored by different agents in disjoint sub-regions which have little overlap \cite{wurm200multirobot}, leading to a set of partial scans. The alignment of such unordered partial 3D data is an under-explored problem and it is challenging to the existing methods because of their requirements on the large overlap and dense feature points for scan registration. 

\begin{figure}[t]  
    \centering
  \begin{overpic}[width=0.95\linewidth]{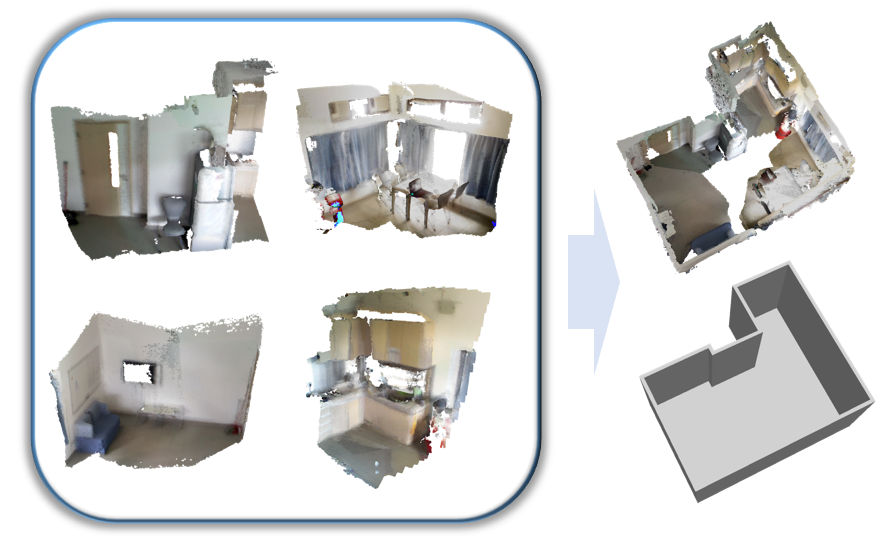}
    \end{overpic}
    \vspace{-4mm}
  \caption{We present a method to jointly align a set of unordered partial reconstructions and estimate a room layout.}
  \label{fig:teaser}
  \vspace{-4mm}
\end{figure} 

In this paper, we propose a method for registering partial reconstructions of an indoor scene which may not have sufficient overlap, as shown in Fig. \ref{fig:teaser}. Our key observation is that the local layouts of partial reconstructions can be viewed as the fragments of a global room layout which typically has the following two characteristics: (1) the room layout is a set of perpendicular or parallel walls, which is referred to the Manhattan World (MW) property; (2) the room layout forms a simple closed loop on a 2D floorplan. We exploit these properties to develop an efficient method for jointly predicting a room layout that has the above layout properties and estimating the transformations from a set of unordered partial reconstructions.

\begin{figure*}
    \centering
    \begin{overpic}[width=\linewidth]{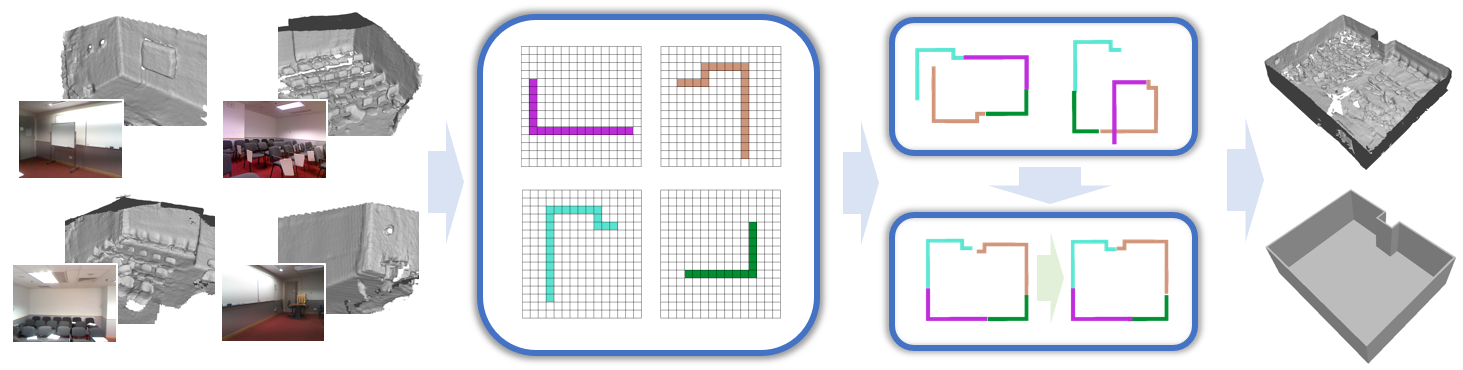}
      \put(40, -5) {\small (a) Input partial data}
      \put(163, -5) {\small (b) Local floorplan estimation}
       \put(315, 126) {\small (c) Global placement}
    \put(280, -5) {\small (d) Optimal placement and refinement}
         \put(410, 126) {\small (e) Partial data alignment}
       \put(425, -5) {\small (f) Layout modeling}
    \end{overpic}
    \vspace{-3mm}   
    \caption{Overview of the proposed method. Given a set of unordered partial reconstructions (a), our algorithm first estimates their local floorplans (b) respectively. Then we compute the poses (c) of all the local floorplans to find a global optimal placement followed by a refinement process (d). Finally, we output the aligned complete reconstruction (e) and predict a final room layout (f) accordingly. }
    \label{fig:overview}
    \vspace{-4mm}   
\end{figure*}
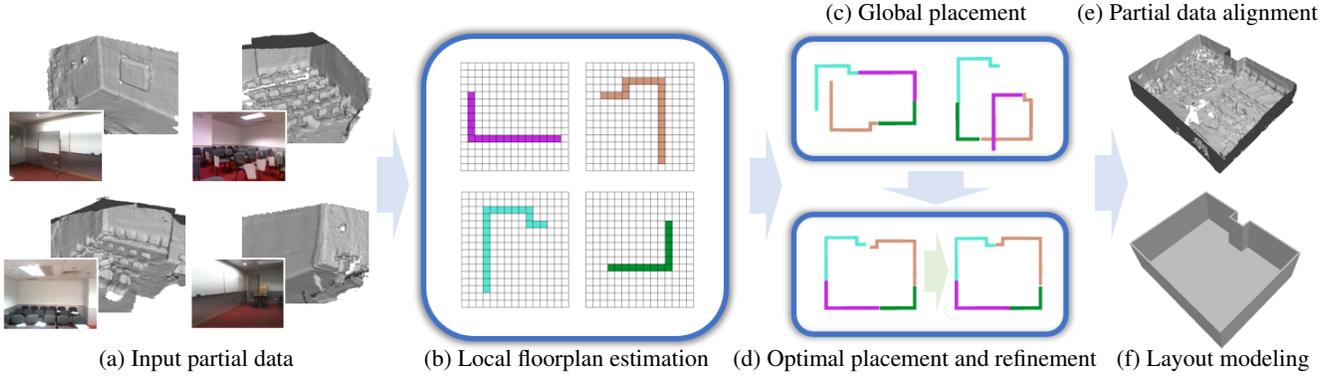

Most of the existing methods \cite{cabral2014piecewise,lee2017joint} use boundary loop detection to estimate a room layout because their input is a long sequence of scans that have substantial overlap and complete coverage of the indoor scene. In contrast, the input to our method can be partially scanned data without clear boundaries. By taking noises and occlusions into consideration, our method is capable of reconstructing scenes with incomplete, disconnected or even occluded walls. Given such a set of partial scans with detected layouts, we analyze the relationship between each local layout with the global layout to achieve successful alignment, while the existing methods would fail due to the lack of sufficient overlap and features for establishing correspondences. We formulate a novel optimal placement problem to determine the rotation and translation of each partial scan using the MW assumption and the layout properties, and then produce the final transformations to align the scans and predict a complete global room layout. The framework of our method is illustrated in Fig. \ref{fig:overview}.

Without relying on feature matching, our method not only works robustly when the partial reconstructions do not have substantial overlap, but also enables a series of novel applications, e.g., the reconstruction of featureless or large scenes, modeling from sparse input, RGB-D stream down-sampling, to name a few (Sec. \ref{sec:application}).

We validate our approach qualitatively and quantitatively on both real and synthetic scenes of various sizes and complexities, and compare it with the state-of-the-art methods. The evaluations and comparisons demonstrate that, given a set of partial reconstructions, our method is able to compute the accurate transformations to align them and reconstruct a high-quality scene layout by effectively estimating and combining local layouts of partial data.

\section{Related Work}
Indoor scene understanding has been a popular topic and accumulated rich literature in the past decades. We review the most relevant works and refer readers to the survey \cite{naseer2018indoor} to have an overview. 

\para{3D data registration}In the last decade, a number of simultaneous localization and mapping (SLAM) techniques are extensively employed to model 3D scenes using RGB-D sensors. Some typical works include Kinect Fusion \cite{newcombe2011kinectfusion}, Elastic Fusion \cite{whelan2016elasticfusion}, ORB-SLAM \cite{mur2017orb} and so on. To establish robust correspondences between 3D data, a wide range of geometric feature descriptors \cite{rusu2009fast, zeng20163dmatch, johnson1999spin} are proposed. Also, global registration approaches \cite{yang2013goicp, zhou2016fast} are developed to alleviate the local optimum issue when aligning point sets. These methods are effective for feature matching, surface alignment as well as 3D reconstruction. However, when it comes to the 3D data without sufficient overlap and correspondences, these algorithms are likely to fail or exhibit unacceptable inaccuracies (see Fig. \ref{fig:quali_pose_cmp} and Fig. \ref{fig:featureless}).

\para{Room layout estimation}Methods for room layout estimation can be roughly divided into three categories based on their inputs, i.e., single view RGB/RGB-D image, panoramic RGB/RGB-D image, and dense point cloud.

Many works focusing on layout estimation from a single image \cite{lee2009geometric, schwing2012efficient,hedau2009recovering, choi2013understanding,schwing2013box} have been continuously developed. Due to the limitation of the narrow field-of-view caused by a single standard image, researchers have tried to exploit panoramic images \cite{zhang2014panocontext, cabral2014piecewise, yang2016efficient} to recover the whole room context. With the success of deep learning in vision tasks, newest techniques \cite{lee2017roomnet, zou2018layoutnet} rely on convolutional neural networks to map an RGB image to a room layout directly. These methods using standard or panoramic RGB images are highly dependent on feature points either for key structure detection or for pose estimation. Because of the instability of image feature points, these methods will suffer from inaccuracy as well as the incapability of handling complex (they usually recover ``cuboid'' or ``L'' shape \cite{lee2017roomnet}) and featureless scenes. Instead, our method uses depth data and is independent of feature points to avoid these drawbacks. 

RGB-D images include 3D range information of each pixel, thus significantly improving the accuracy and the robustness of geometry reasoning. Some methods use a single RGB-D image \cite{taylor2013parsing, zhang2013estimating} to estimate room layout, which is also limited by the narrow field-of-view. With the superiority of panoramic RGB-D images, higher-quality layout analysis and structured modeling results have been achieved \cite{ikehata2015structured, wijmans2017exploiting}. There are also a few methods using densely scanned point clouds as input to estimate scene layouts \cite{murali2017indoor,lee2017joint,liu2018floornet}. Most of these methods target a complete scene in order to exploit the closed boundary nature of room layout, while our method is able to cope with the more challenging partial scans which lack clear outer boundaries.

\para{Indoor scene constraints}Intrinsic properties of indoor scenes are widely used in indoor understanding and reconstruction. Manhattan World (MW) assumption is the predominant rule, thus Manhattan frame estimation is well researched for both RGB \cite{lee2009geometric,schwing2012efficient} and RGB-D images \cite{ghanem2015robustmanha,joo2016globally}. MW assumption serves as a guidance in many applications such as layout estimation \cite{lee2009geometric, schwing2012efficient,hedau2009recovering, choi2013understanding,schwing2013box, yang2016efficient}, camera pose estimation \cite{straub2015realrotate, kim2017visual} and reconstruction refinement \cite{halber2017fine, huang20173dlite}. 

In addition to the MW assumption, indoor scenes have plentiful lines and planes which provide strong cues for many tasks. Elqursh and Elgammal \cite{elqursh2011line}  introduce a line-based camera pose estimation method, while Koch \etal \cite{koch2016alinelline} use 3D line segments to align the non-overlapping indoor and outdoor reconstructions. Planar patch detection and matching \cite{taguchi2013pointplane, ma2016cpa_planemodealign, concha2015dpptam_plane1, salas2014dense_plane2, shi2018planematch, halber2017fine, lee2017joint} are significantly used strategies to improve the reconstruction accuracy. Some works \cite{taguchi2013pointplane, ma2016cpa_planemodealign, concha2015dpptam_plane1, salas2014dense_plane2} exploit plane correspondence to solve for frame-to-frame camera poses. Halber \etal \cite{halber2017fine} and Lee \etal \cite{lee2017joint} perform global registration leveraging structural constraints to elevate the scan accuracy. Shi \etal \cite{shi2018planematch} use a CNN to learn a feature descriptor for planar patches in RGB-D images.  
These approaches all hinge on the success of feature matching at the overlapping areas, as opposed to the scenario in this paper. 

\section{Approach}
\label{sec:approach}
The input to our system is a set of partially scanned fragments and we output the local layout of each fragment, the transformations to align them, and a global scene layout. As shown in Fig. \ref{fig:overview}, our approach consists of three main steps: (1) local layout estimation of each partial reconstruction; (2) optimal placement for global layout estimation; (3) pose refinement to make walls well-aligned. Before running our algorithm, we first extract point feature \cite{rusu2009fast} to combine the partial scans that have more than 60$\%$ alignment inliers into one fragment; while the remaining scans can be considered as insufficiently overlapping.

\subsection{Local Layout Estimation}
\label{sec:localEstimate_pre}
We assume that walls obey the MW assumption. Inspired by Cabral and Furukawa \cite{cabral2014piecewise}, we formulate a graph-based shortest path problem to find a floorplan path. As opposed to their reliance on a complete point cloud with a closed-loop as input, we come up with new strategies dealing with partial input that may contain incomplete or partially occluded walls.

\para{Preprocessing}We extract the planes using RANSAC and compute three MW directions $\{ X_m, Y_m, Z_m\}$ \cite{joo2016globally}. For convenience, we set the $X_m$ axis as the world up direction by assuming that the camera optical axis is roughly horizontal to the ground when the scanning begins, and the $Y_m$ and $Z_m$ axes are the wall directions. Then the local camera coordinates are aligned to the MW coordinates by the minimal rotation.

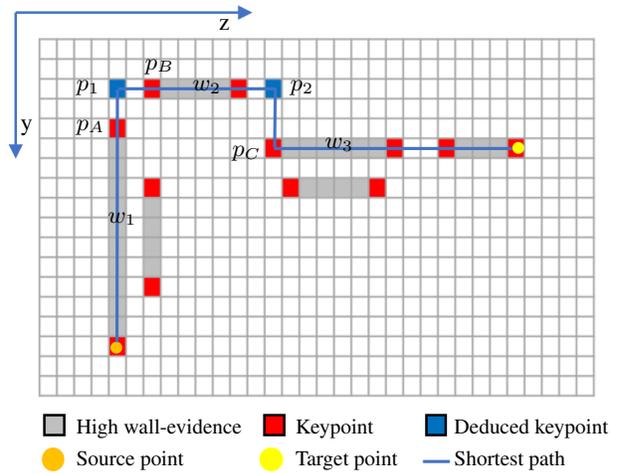
\begin{figure}[h]
  \begin{overpic}[width=0.95\linewidth]{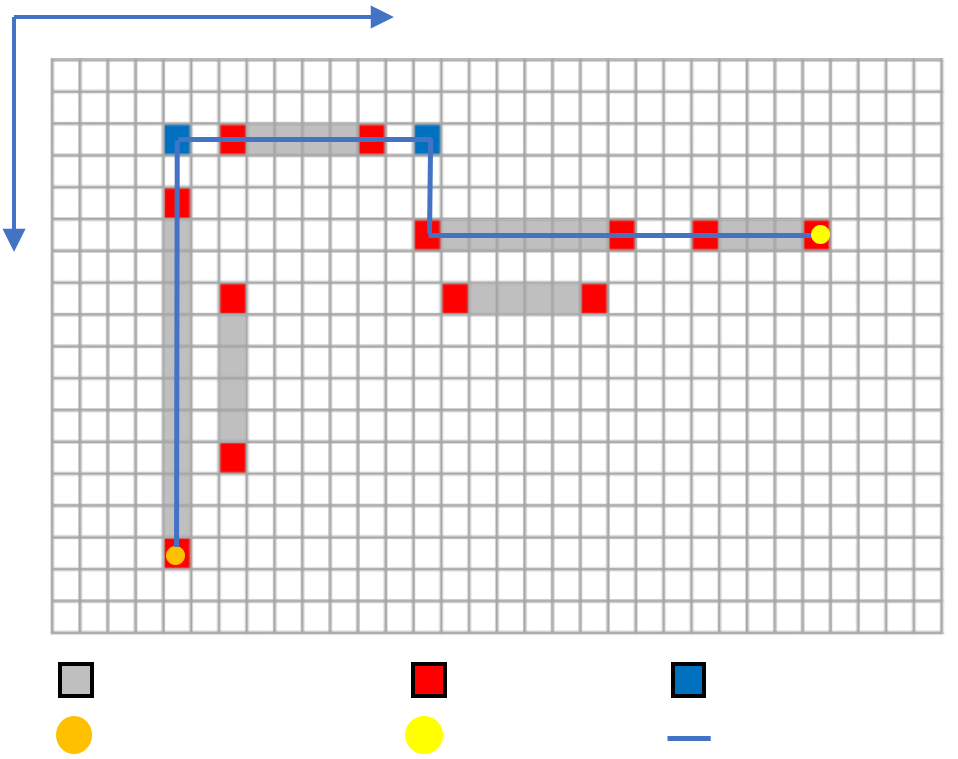}
     \put(26, 15) {\footnotesize High wall-evidence}
        \put(109, 15) {\footnotesize Keypoint}
        \put(169, 15) {\footnotesize Deduced keypoint}
        \put(26, 3) {\footnotesize Source point}
        \put(109, 3) {\footnotesize Target point}
        \put(169, 3) {\footnotesize Shortest path}
        \put(5, 130) {\small y}
        \put(80, 168) {\small z}
        \put(26, 130) {\small $p_A$}
        \put(26, 145) {\small $p_1$}
        \put(52, 153) {\small $p_B$}
        \put(107, 145) {\small $p_2$}
        \put(85, 120) {\small $p_C$}
        \put(38, 95) {\small $w_1$}
        \put(70, 145) {\small $w_2$}
        \put(120, 123) {\small $w_3$}
    \end{overpic}
    \vspace{-2mm}
  \caption{Local floorplan path determination. Points are projected onto the ground plane and discretized into a grid.}
  \label{fig:wall}
\vspace{-6mm}
\end{figure}

\para{Wall estimation graph}We project all points of the fragment $f_k$ onto a grid with cell size $s$. A cell that receives more than $N$ projected vertices is considered as a high wall-evidence cell, where we use $s=8cm$ and $N=20$ in this paper. We search over the grid to look for contiguous sets of cells with high wall-evidence to extract candidate wall segments, such as $w_1, w_2$ and $w_3$ in Fig. \ref{fig:wall}. 

Given a set of wall candidates, we build a wall estimation graph (\emph{WE-graph}) where the nodes are the candidate keypoints of wall structures (e.g., wall corners) and the edges are the candidate walls. Due to noise and occlusion, the endpoints (red cells in Fig. \ref{fig:wall}) may not exactly be wall corners. We therefore need to reason out more candidate keypoints (e.g., $p_1, p_2$) to derive a complete wall structure. 

Here we consider two typical cases: (1) two neighboring perpendicular candidate wall segments can be extended to an intersection point which may imply a potential wall corner, e.g., $p_1$ is deduced from $w_1$ and $w_2$ in Fig. \ref{fig:wall}; (2) two neighboring misaligned parallel candidate wall segments may imply an occluded wall in the invisible intermediate region. See $w_2$ and $w_3$ in Fig. \ref{fig:wall}, we project $p_C\in w_3$ to the line of $w_2$ to deduce a new keypoint $p_2$, and re-mark the cells between $p_2$ and $p_C$ as high wall-evidence.

We set both of the deduced points (blue cells) and the wall endpoints (red cells) as the graph nodes. Then edges are added for every pair of the nodes as long as they are aligned to either $Y_m$ or $Z_m$ axis. The edge weight of a potential wall $w$ is defined as
\begin{equation}
\setlength{\abovedisplayskip}{10pt} 
\setlength{\belowdisplayskip}{5pt} 
\frac{L(w)-H(w)}{H(w)}+\lambda \text{,}
\label{eq:wallestimate}
\end{equation}
where $L(w)$ is the length of $w$ on the grid, and $H(w)$ is the number of high-evidence cells. The first term is to encourage edges to not only have fewer low wall-evidence cells but also be longer. The second term is a constant complexity penalty with $\lambda=0.1$ (see the evaluation in Fig. \ref{fig:lambda}). Through these two terms, we encourage the final path to have higher wall-evidence, be longer and simpler.

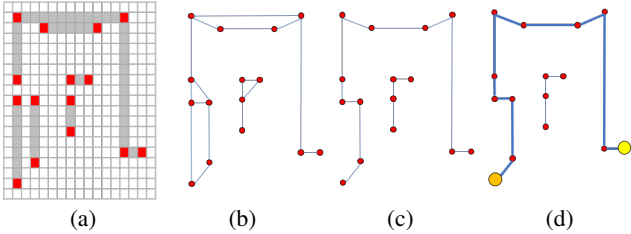
\begin{figure}[h]
\begin{overpic}[width=\linewidth]{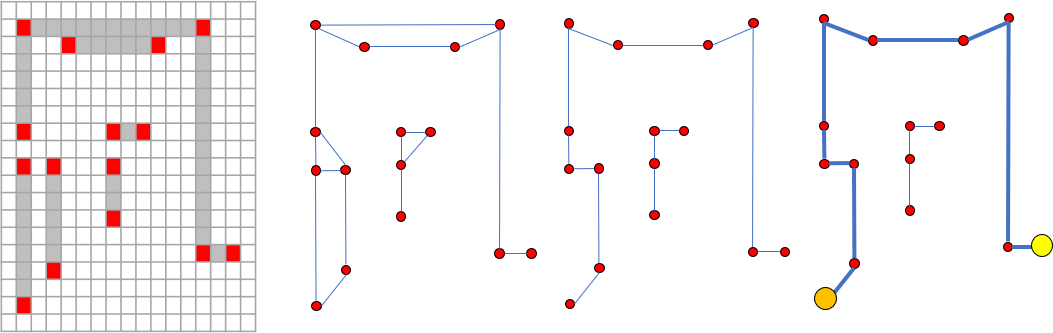}
     \put(25, -10) {\small (a)}
        \put(85, -10) {\small (b)}
        \put(145, -10) {\small (c)}
        \put(205, -10) {\small (d)}
    \end{overpic}
    \vspace{-3mm}
  \caption{Source and target point determination for a partial scan. (a) Projection grid; (b) ST-graph; (c) minimal spanning forest (MSF); (d) source and target points derived from the longest path on the MSF.}
  \label{fig:tree}
\vspace{-2mm}
\end{figure}
\para{Source and target determination} 
To solve for the floorplan path from an incomplete reconstruction that does not have a clear boundary, as shown in Fig. \ref{fig:tree}, we build another graph (\emph{ST-graph}) to determine the source and target points. The edge weight in the \emph{ST-graph} is the Euclidean distance between two nodes in the grid coordinate system. We compute the minimal spanning forest (MSF) of the graph to encourage the nodes to be connected by the minimal distance cost. Then we solve for the longest path on the MSF. The source and the target points are two endpoints of this longest path, where the first point in the clockwise sequence is considered as the source and the other as the target. 

Finally, we find the minimum cost path from the source to the target on the \emph{WE-graph} as the local layout estimation result.

\subsection{Global Layout Placement}
\label{sec:global_place}
\begin{figure}[!htb]
 \centering  
 \vspace{-6mm}
 \begin{overpic}[width=\linewidth]{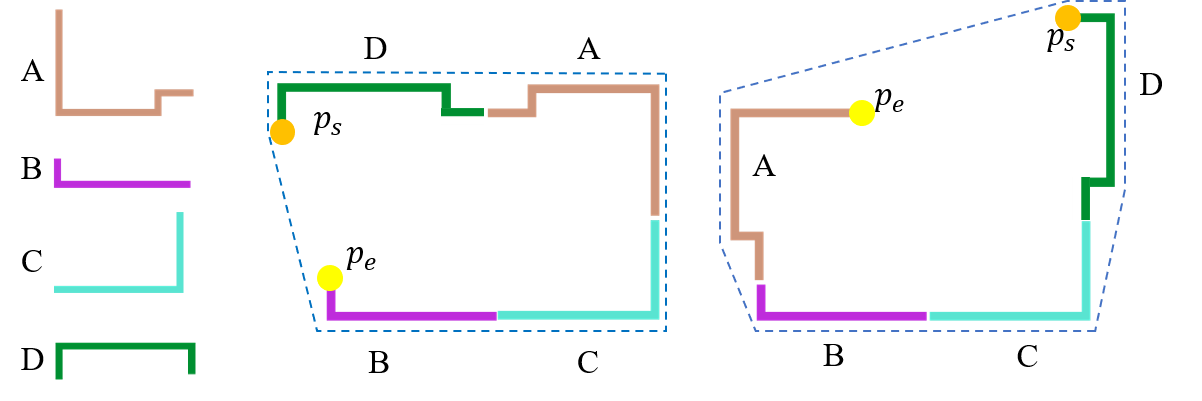}
        \put(95, -5) {\small (a)}
        \put(180, -5) {\small (b)}
    \end{overpic}
    \vspace{-4mm}
  \caption{Two different placements via end-to-end local layout concatenation. }
  \label{fig:convexhull}
\vspace{-3mm}
\end{figure}

To determine the global layout, we need to find the rigid transformations for all partial fragments that do not have sufficient matched-overlap. We observe that under the MW assumption, the rotation of each partial fragment can be viewed as the alignment of its local MW coordinate to the world one; the translations of the small-overlapping fragments can be approximately viewed as the sequence in the global loop closure path where all of the local paths are concatenated end-to-end, see Fig. \ref{fig:convexhull} for an example.

Given the local MW coordinate axes $\{X_m,Y_m,Z_m\}$ of a fragment and the world coordinate axes $\{X_w, Y_w, Z_w\}$, we first align the up direction $X_m$ of the local MW coordinate to the world up direction $X_w$ (see Preprocessing in Sec. \ref{sec:localEstimate_pre}) . Then the remaining correspondences from $Y_m, Z_m$ to $Y_w, Z_w$ have four different choices which compose the solution space of rotations. Let $f\in\{1,...,N\}$ index all the partial fragments, $R_f \in \{1,2,3,4\}$ the candidate rotations of fragment $f$ corresponding to the alignment from $Y_m$ to $Y_w$, $Y_m$ to $-Y_w$, $Y_m$ to $Z_w$ or $Y_m$ to $-Z_w$ respectively, and $t_f \in \{1,...,N\}$ the clockwise sequence of the fragment $f$ on the floorplan loop.

A candidate placement is denoted as a tuple $\{f, R,t\}$ where the subscript is omitted for simplicity. It indicates the rotations and sequences for all the fragments as well as the room layout derived by the end-to-end concatenation of the local layout paths. We then define the binary variables $x_{f,R,t}\in\{0,1\}$ to indicate whether the candidate placement exists in the solution set.
The total energy is defined as
\begin{equation} \label{eq:plc_obj}
\setlength{\abovedisplayskip}{5pt} 
\setlength{\belowdisplayskip}{10pt} 
\min\limits_{\bm{x}=\{x_{f,R,t}\}} E_l(\bm{x}) +E_c(\bm{x})+ E_b(\bm{x}) \text{,}
\end{equation}
\begin{equation} \label{eq:plc_cons}
\setlength{\abovedisplayskip}{5pt} 
\setlength{\belowdisplayskip}{10pt} 
s.t. \quad \forall f \sum\limits_{R,t}x_{f,R,t} = 1\text{,}  \quad \forall t \sum\limits_{f,R}x_{f,R,t} = 1\text{,}
\end{equation}
where $E_l$ is the complexity of a layout, $E_c$ the closure measurement, and $E_b$ the similarity of the boundary between adjacent fragments. The constraints in Eq. (\ref{eq:plc_cons}) enforce mutual exclusion, i.e., each fragment and sequence index can only appear once in the final solution. 

\para{Layout complexity term}We form the complexity term $E_l$ by summing up the number of wall corners and the number of edges in the convex hull of the floorplan, where the lowest energy encourages that the room not only contains fewer corners but also has simpler overall structure. See Fig. \ref{fig:convexhull}, (a) and (b) are two different placements for the same set of local layouts. Although they have the same number of wall corners, we prefer (a) since it has more aligned collinear wall segments which lead to fewer edges in the convex hull.

\para{Closure term}The second term $E_c$ denotes the closure of a layout path, by which we wish the gap between the start point and the endpoint on the final path to be as small as possible. See Fig. \ref{fig:convexhull} for an example of computing this term, the closure is measured by the Manhattan distance (in meters) between the start point $p_s$ of $x_{f,R,1}$ and the endpoint $p_e$ of $x_{f,R,N}$.

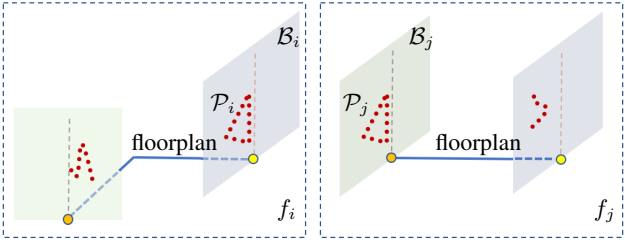
\begin{figure}
\centering
 \begin{overpic}[width=\linewidth]{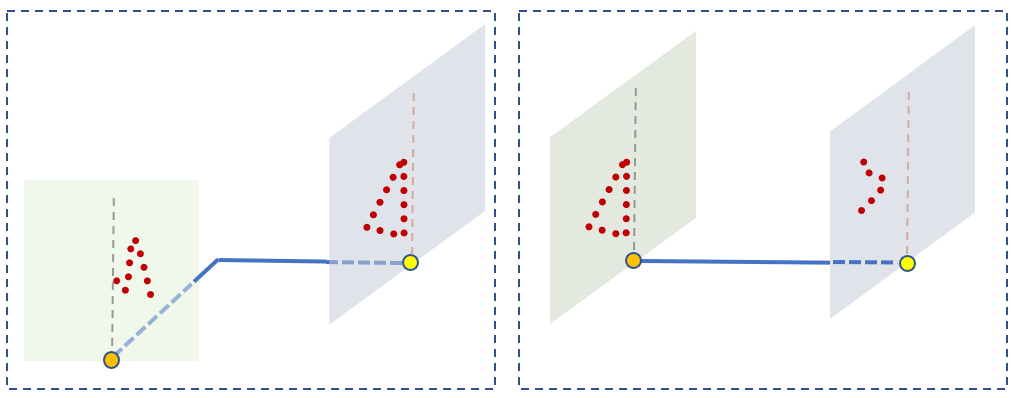}
         \put(105, 75) {\small $\mathcal{B}_i$}
         \put(155, 75) {\small $\mathcal{B}_j$}
         \put(80, 50) {\small $\mathcal{P}_i$}
         \put(130,50) {\small $\mathcal{P}_j$}
          \put(105, 10) {\small $f_i$}
         \put(225,10) {\small $f_j$}
         \put(50,35) {\small floorplan}
         \put(165,35) {\small floorplan}
    \end{overpic}
    \vspace{-4mm}
  \caption{Analysis of the boundary similarity when $f_j$ is placed next to $f_i$.  $\mathcal{B}_i$ and $\mathcal{B}_j$ are two adjacent boundary planes; $\mathcal{P}_i$ and $\mathcal{P}_j$ are the boundary point sets around the planes, which are used for computing boundary similarity.}
  \label{fig:boudnarysim}
\vspace{-4mm}
\end{figure}

\para{Boundary similarity term}As shown in Fig. \ref{fig:boudnarysim}, the cutting plane going through the source or the target point on a local floorplan path is defined as the boundary plane (e.g., $\mathcal{B}_i$ and $\mathcal{B}_j$). The points within $10cm$ of the cutting plane are considered as the boundary points (e.g., $\mathcal{P}_i$ and $\mathcal{P}_j$).  We refer to the probabilistic method \cite{bogoslavskyi2017analyzing} to analyze the match quality of the boundary points between two adjacent fragments, and obtain a mismatch score between 0 and 1. We sum up the mismatch scores of all adjacent pairs to compute $E_b$.


To solve this constrained 0-1 programming problem (Eq.~(\ref{eq:plc_obj})), we search for the global minima based on a DFS tree with alpha-beta pruning. Additionally, we also prune the invalid branches where walls incorrectly cross each other to further improve the efficiency.

\subsection{Pose Refinement}
The global layout placement encourages all fragments to form a loop closure without taking wall alignment into consideration. Thus in this step, we aim to refine the positions of all fragments by constraining the layout alignment.

\begin{figure}[!htb]
\centering
  \vspace{-2mm}
 \begin{overpic}[width=\linewidth]{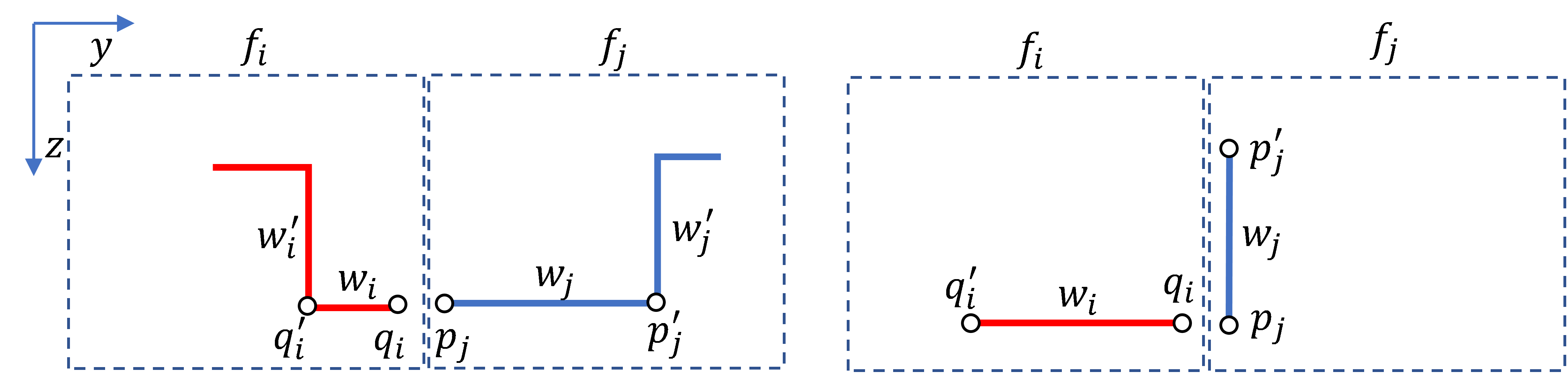}
        \put(60, -10) {\small (a)}
        \put(175, -10) {\small (b)}
    \end{overpic}
    \vspace{-3mm}
  \caption{Two types of wall joints between two adjacent fragments $f_i$ and $f_j$. (a) The connected walls are parallel; (b) the connected walls are perpendicular.}
  \label{fig:connect}
\vspace{-2mm}
\end{figure}

\begin{figure*}
\centering
\vspace{-2mm} 
 \begin{overpic}[width=\linewidth]{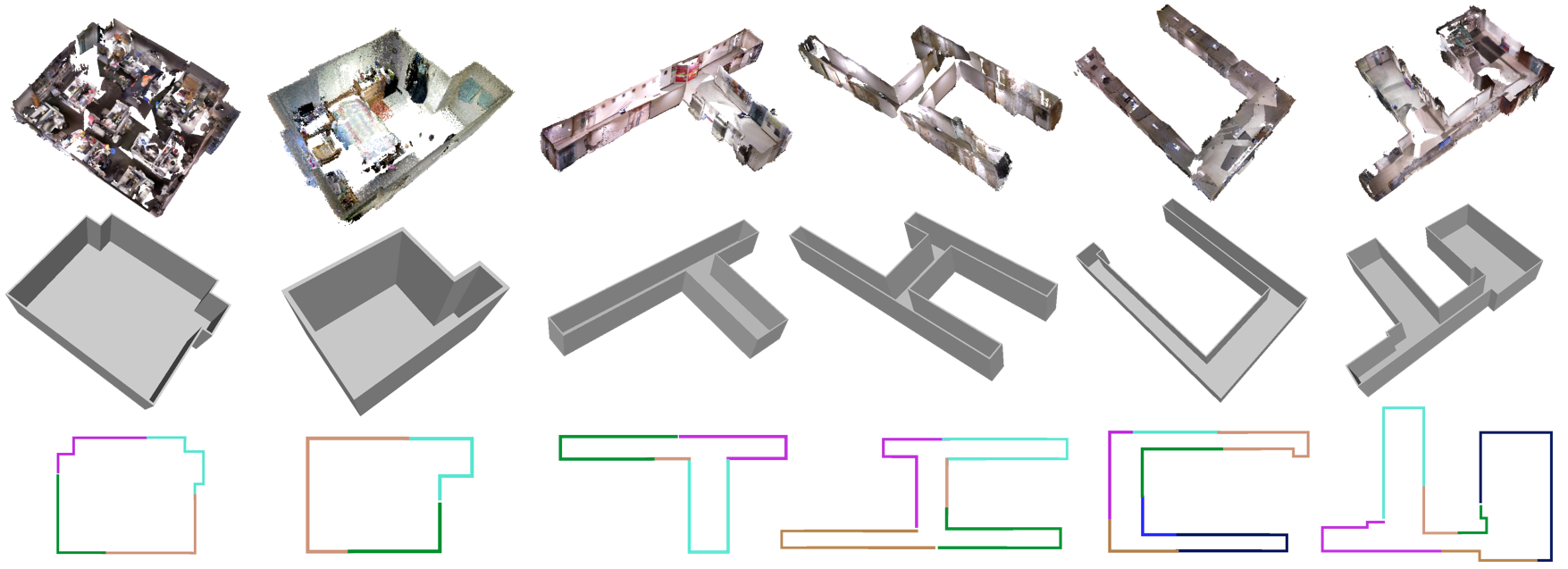}
    \end{overpic}
    \vspace{-5mm} 
  \caption{Results of the partial reconstruction alignment and the global layout estimation. }
  \label{fig:global_results}
\vspace{-3mm}
\end{figure*}

Let the sequence of local layouts be $\{f_1,f_2,...,f_N\}$ on the loop. Since the walls are aligned to either $Y$ or $Z$ axis of the world coordinate system, we define $t_i=(y_i,z_i)$ to represent the translation to adjust the current position of the layout $f_i$. Meanwhile, we use $q_i$ and $p_j$ to denote the target point in $f_i$ and the source point in $f_j$ respectively, while $p_i'$ and $q_j'$ are their neighboring keypoints (corner-point or end-point) in the same local layout accordingly (see Fig. \ref{fig:connect} for an illustration). There are two typical configurations of wall connection when $f_j$ is placed next to $f_i$ and the constraints are added accordingly as follows.

\para{Parallel connection (Fig. \ref{fig:connect} (a))}Two adjacent local layouts $f_i$ and $f_j$ are joined by two parallel walls. The walls are aligned along either the $Y$ axis or the $Z$ axis, while we only discuss the $Y$-aligned case which is shown in Fig. \ref{fig:connect} (a).
First, the $Z$ coordinates of $q_i$ and $p_j$ should be equal or else the walls are misaligned. Second, given two joined walls $w_i$ and $w_j$ with the lengths $l_{w_i}$ and $l_{w_j}$ respectively, if $l_{w_i} \le l_{w_j}$, then $p_j$ can not go across $q_i'$ or else $w_j$ will intersect with $w_i'$ which is illegal. The constraints are defined as follows where $\alpha=min\{ l_{w_i},l_{w_j}\}$:
\begin{equation}
\setlength{\abovedisplayskip}{5pt} 
\setlength{\belowdisplayskip}{5pt} 
\begin{split}
& z_{q_i}+z_i = z_{p_j}+z_j, \\
& (l_{w_i}+l_{w_j})-|(y_{q_i'}+y_i)-(y_{p_j'}+y_j)|<\alpha.
\label{eq:refine_cons1}
\end{split}
\end{equation}

\para{Perpendicular connection (Fig. \ref{fig:connect} (b))}Two adjacent local layouts $f_i$ and $f_j$ are jointed by two perpendicular walls. We only discuss the case of Fig. \ref{fig:connect} (b) where $w_i$ is aligned along the $Y$ axis and $w_j$ the $Z$ axis.
To avoid illegal crossing between $w_i$ and $w_j$, $p_j$ cannot go across $w_i$ while $q_i$ cannot go across $w_j$. The constraints are defined as:
\begin{equation}
\setlength{\abovedisplayskip}{5pt} 
\setlength{\belowdisplayskip}{5pt} 
\begin{split}
& y_{q_i}+y_i<y_{p_j}+y_j\\
& z_{p_j}+z_j<z_{q_i}+z_i.
\end{split}
\label{eq:refine_cons2}
\end{equation}

To solve for the adjustments $\bm{t}=\{(y_i,z_i)\}$ for all pairs of local layouts, we formulate an optimization problem to minimize the distance between the joints of the adjacent local layouts as follows: 
\begin{equation} \label{eq:refine}
\setlength{\abovedisplayskip}{5pt} 
\setlength{\belowdisplayskip}{5pt} 
\min\limits_{\bm{t}} \sum_{(i,j)\in \mathcal{C}} ((q_i+t_i)-(p_j+t_j))^2.
\end{equation}
Here $\mathcal{C}$ indicates the set of the pairs of the adjacent local layouts. Finally, we obtain the translations $\{(y_i,z_i)\}$ for all local layouts by solving Eq.~(\ref{eq:refine}) under the constraints (\ref{eq:refine_cons1}) and (\ref{eq:refine_cons2}), and update the final layout.

\section{Experimental Results}
We evaluate our algorithm using 101 scenes collected from SUNCG dataset \cite{song2017ssc}, SUN3D dataset \cite{Xiao_2013_ICCV} and our real-world scanning. Each scene is given by a set of partial reconstructions derived from the region-by-region capturing or the failures of camera localization. A challenge in our testing data is, there may not be sufficient overlap among the partial data.
Our dataset covers representative indoor layouts of which the scene area varies from $2m\times 6m$ to $18m \times 20m$, and the number of wall corners varies from 4 to 16. All the experiments are performed on a machine with Intel Core i7-7700K 4.2GHz CPU and 32GB RAM.

\para{Evaluation metrics}We evaluate the performance of our method by the metrics defined below. A local or global layout estimation is correct if the average distance error between the estimated wall keypoints and the ground truth keypoints is below 5\% relative to the length of the diagonal of the bounding box. A global placement is correct if the placement can lead to a correct global layout estimation. We use $ACC_{local}$ to represent the percentage of the correct local estimations against all of the partial fragments in the dataset. Similarly, $ACC_{global}$ represents the percentage of the correct global placements against all scenes.

\begin{figure}[!htb]
\centering
\vspace{-2mm}
 \begin{overpic}[width=0.95\linewidth]{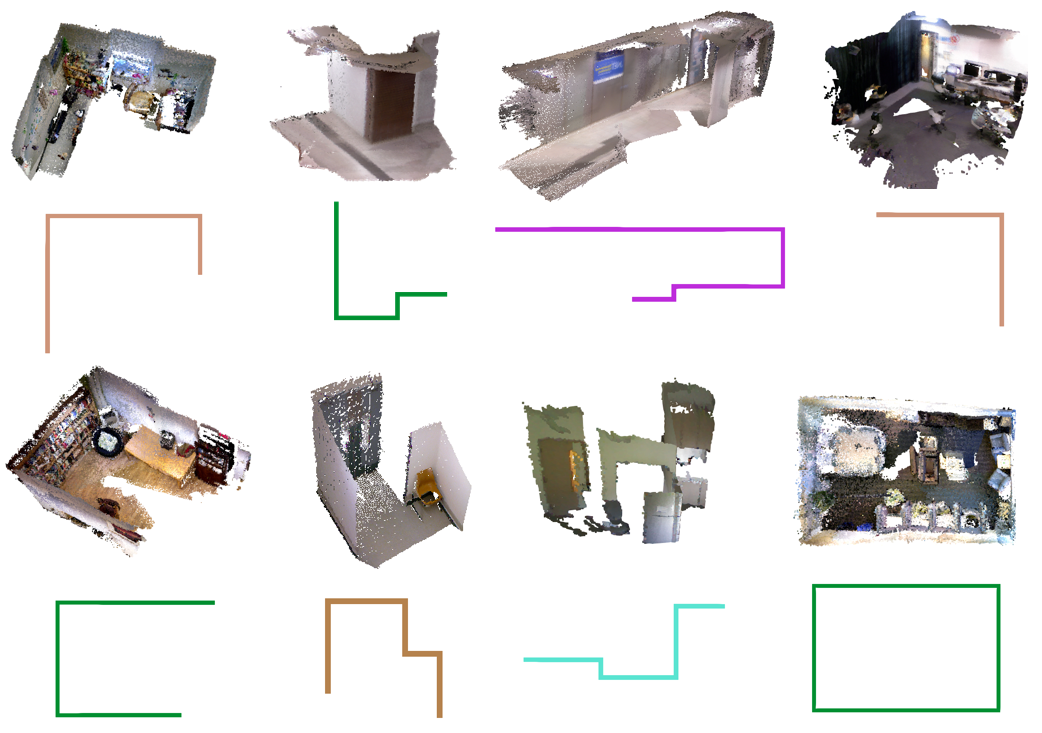}
    \end{overpic}
       \vspace{-2mm}
  \caption{Results of the partial layout estimation.}
  \label{fig:partial_results}
   \vspace{-2mm}
\end{figure}

\para{Partial layout estimation}Our method is able to robustly estimate a partial room layout given an incomplete reconstruction without a closed boundary. Our testing data contains 401 various partial reconstructions, on which our method achieves $ACC_{local}=98.3\%$. We also show some qualitative results in Fig. \ref{fig:partial_results}.  Note that: (1) some walls are not captured in the point cloud but our method can still robustly estimate the correct layouts; (2) although our method targets partial data, it can be directly applied to estimate the layout of a complete reconstruction as well.

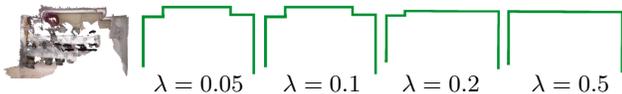
\begin{figure}[!htb]
\begin{overpic}[width=\linewidth]{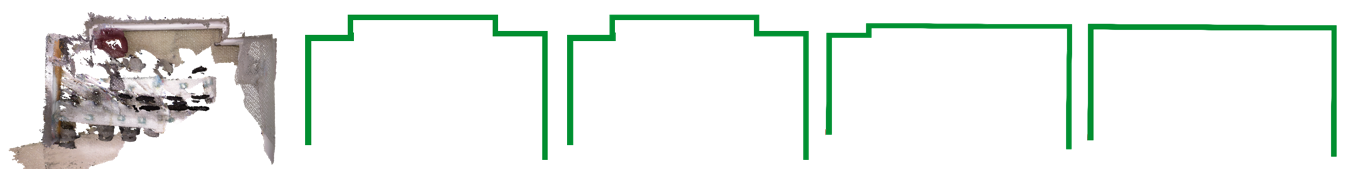}
\put(58, -5) {\small $\lambda=0.05$}
\put(107, -5) {\small $\lambda=0.1$}
\put(152, -5) {\small $\lambda=0.2$}
\put(201, -5) {\small $\lambda=0.5$}
\end{overpic}
  \caption{The effect of the parameter $\lambda$ of the penalty term.}
  \label{fig:lambda}
 \vspace{-2mm}
\end{figure}

We evaluate the effect of different values of the complexity penalty $\lambda$ in Eq. (\ref{eq:wallestimate}). Fig. \ref{fig:lambda} shows that a large $\lambda$ tends to ignore the detailed structures and produce a simple layout. We fix $\lambda$ to 0.1 to generate all of the results in this paper.

\para{Global layout placement}Fig. \ref{fig:global_results} shows some results of the partial scan alignment and the global layout estimation. Our method faithfully reconstructs some large-scale scenes by combining a set of partially scanned point clouds. We also quantitatively evaluate our method in Table \ref{tab:ablation_study}. As an ablation study, Table \ref{tab:ablation_study} shows the performance given different configurations of the three terms in Eq. (\ref{eq:plc_obj}): (1) without closure term; (2) without complexity term; (3) without boundary similarity term; (4) full terms. The experiments demonstrate that the full configuration using all these three terms performs the best.

\begin{table}[!htb]
\centering
\begin{tabular}{cc}
\hline
Configuration                    & $ACC_{global}$(\%)  \\\hline
w/o closure term             & 22.8           \\
w/o complexity term          & 67.5          \\
w/o boundary similarity term & 80.2          \\
full terms                        & \textbf{85.1} \\\hline
\end{tabular}
\vspace{-2mm}
\caption{Performance of our method on global layout placement using different configurations.}
\label{tab:ablation_study}
\vspace{-2mm}
\end{table}

\begin{figure}[t]
\begin{overpic}[width=\linewidth]{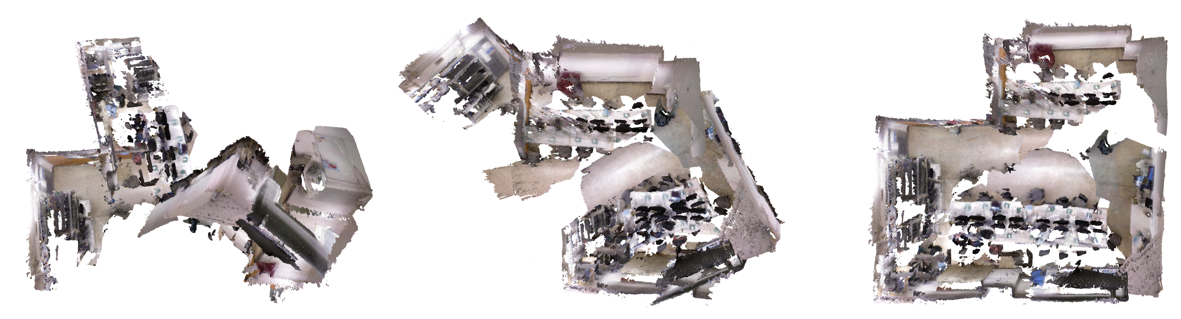}
\put(32, -10) {\small (a) }
\put(118, -10) {\small (b)}
\put(200, -10) {\small (c)}
\end{overpic}
\vspace{-2mm}
    \caption{Qualitative comparison with point cloud alignment methods using feature descriptors. (a) 3DMatch \cite{zeng20163dmatch}; (b) FPFH \cite{rusu2009fast}; (c) ours.}
  \label{fig:quali_pose_cmp}
  \vspace{-2mm}
\end{figure}

\para{Pose estimation error}We evaluate the pose estimation error on the synthetic scenes collected from SUNCG \cite{song2017ssc} dataset with ground truth camera poses. We also compare our method with the state-of-the-art 3D registration algorithms, including 3DMatch \cite{zeng20163dmatch}, Fast point feature histogram (FPFH) \cite{rusu2009fast}, and Orthogonal plane-based visual odometry (OPVO) \cite{kim2017visual}. Note that OPVO is also proposed under the MW assumption. Table \ref{tab:quant_pose_cmp} reports the angle error of rotation and the distance error of translation relative to the length of the diagonal of the bounding box. Since our testing data may not have sufficient overlap, we find that existing methods based on feature descriptors perform poorly under the same condition.  Qualitative comparisons in Fig. \ref{fig:quali_pose_cmp} and quantitative comparisons in Table \ref{tab:quant_pose_cmp} both show that the other methods produce unacceptable inaccuracies, while our method achieves superior results.

\begin{table}[t]
\centering
\begin{tabular}{ccccc}\hline
Method      &Rotation($^\circ$)\quad     & Translation($\%$)\quad     \\ \hline
3DMatch \cite{zeng20163dmatch} & 43.41 & 21.82 \\
FPFH \cite{rusu2009fast} & 40.05 & 29.12 \\
OPVO \cite{kim2017visual} & 43.06 & 20.04 \\
Ours & \textbf{8.79} & \textbf{9.15} \\\hline
\end{tabular}
\vspace{-2mm}
\caption{Quantitative comparison on the SUNCG synthetic dataset \cite{song2017ssc} in terms of rotation angle error and translation distance error.}
\label{tab:quant_pose_cmp}
\vspace{-2mm}
\end{table}

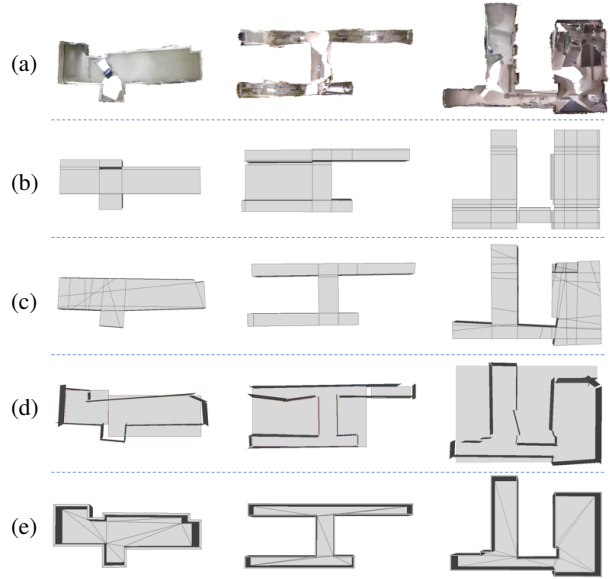
\begin{figure}[!htb]
\centering
\begin{overpic}[width=0.95\linewidth]{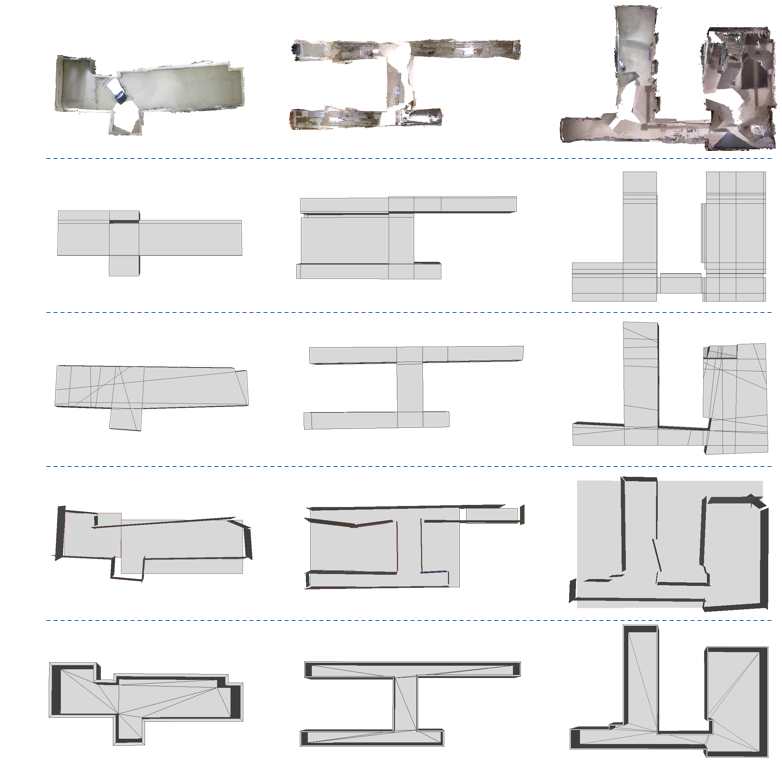}
         \put(-2, 195) {\small {(a)}}
        \put(-2, 150) {\small{(b)}}
        \put(-2, 105) {\small{(c)}}
        \put(-2, 65) {\small{(d)}}
        \put(-2, 20) {\small{(e)}}
    \end{overpic}
 \vspace{-2mm}
\caption{Qualitative comparison on layout reconstruction quality. (a) Input point clouds; (b) MW Modeler \cite{li2016manhattan}; (c) PolyFit \cite{nan2017polyfit}; (d) RAPTER \cite{monszpart2015rapter}; (e) ours.}
\label{fig:cmp}
 \vspace{-4mm}
\end{figure}

\para{Layout reconstruction quality}Manhattan-world Modeler \cite{nan2017polyfit}, PolyFit \cite{li2016manhattan} and RAPTER \cite{monszpart2015rapter} are the state-of-the-art structured modeling methods for man-made scenes which take as input scanned point clouds. To compare with them in terms of layout reconstruction quality, we input to these methods the complete point clouds of the scenes in our dataset. Fig. \ref{fig:cmp} shows a set of qualitative comparison results. We are able to obtain considerably better results with accurate and high-quality wall structures.

\begin{table}[t]
\centering
\begin{tabular}{ccc}
\hline
Method                            & \quad Avg (\%)\quad & \quad Max(\%)\quad \\\hline
MW Modeler \cite{li2016manhattan} & 1.22     & 4.47    \\
PolyFit \cite{nan2017polyfit}     & 1.31     & 5.01    \\
RAPTER \cite{monszpart2015rapter} & 1.40     & 7.84    \\
Ours                      & \textbf{0.90}     & \textbf{2.57}    \\\hline
\end{tabular}
\vspace{-2mm}
\caption{Comparison with the state-of-the-art structured modeling methods in terms of layout reconstruction error.}
\label{tab:quantcmp}
 \vspace{-4mm}
\end{table}

Table \ref{tab:quantcmp} shows the quantitative comparison results with these methods.  We uniformly sample points on the ground truth layout, and compute the distance error of the point samples to their nearest faces in the reconstructed model. We report the average and maximal error relative to the length of the diagonal of the bounding box. The results demonstrate that our method has smaller layout reconstruction errors than the other structured modeling methods. 

\para{Time efficiency}For the local layout estimation, on average our algorithm takes about 0.1s per 10k points. An exception is the scene of the last column in Fig. \ref{fig:global_results}, where it takes about 200s to process a partial scan with 200k points. This is because a large number of small wall candidates are generated in the local layout estimation step due to heavy noises. For the pose determination and refinement, it takes less than 20s with an input of fewer than 10 fragments.

\begin{figure}[!htb]
\vspace{-4mm}
\centering
    \begin{overpic}[width=\linewidth]{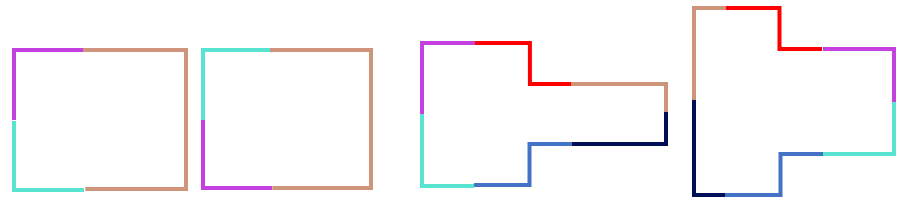}
        \put(55, -10) {\small (a)}
        \put(175, -10) {\small (b)}
    \end{overpic}
    \vspace{-2mm}
\caption{Ambiguity of placements. (a) Different placements produce the same layout; (b) different placements produce different layouts but both are reasonable.}
\label{fig:limit2}
    \vspace{-4mm}
\end{figure}

\para{Ambiguity and failure case}The optimal placement of the given local layouts may be ambiguous, which will result in an incorrect sequence (Fig. \ref{fig:limit2} (a)) or an incorrect layout (Fig. \ref{fig:limit2} (b)), although all the different results seem to be reasonable. The boundary similarity term in Sec.~\ref{sec:global_place} is designed to alleviate this problem, however, if an ambiguity still occurs, more constraints need to be added to derive the correct result, e.g., user-specific fragment sequence.

\begin{figure}[!htb]
\centering
\begin{overpic}[width=\linewidth]{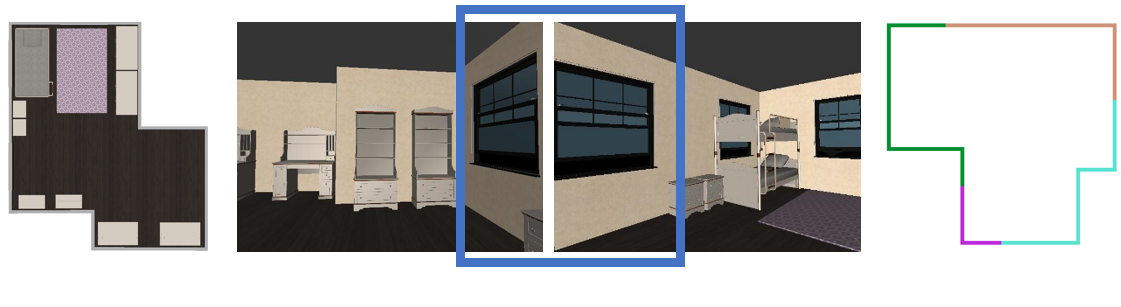}
 \put(0, -4) {\small Ground truth}
 \put(55, -4) {\small Overlap is large but not detected}
  \put(185, -4) {\small Estimated layout}
\end{overpic}
\caption{A failure case where the input fragments have large overlap but not successfully detected by feature descriptor matching.}
\label{fig:limit1}
\vspace{-4mm}
\end{figure}

Before running our algorithm, we first extract point feature \cite{rusu2009fast} to combine the partial scans that have sufficient overlap into larger fragments. If there is large overlap between partial reconstructions but not detected successfully, our algorithm is likely to exhibit large error or output an incorrect result. We show a failure case in Fig. \ref{fig:limit1}, where our result is not consistent with the ground truth.

\section{Applications}
\label{sec:application}
Since our method does not depend on feature matching to align 3D data, it facilitates several novel applications. In this section, we demonstrate the following three.

\begin{figure}[!htb]
\centering
    \begin{overpic}[width=\linewidth]{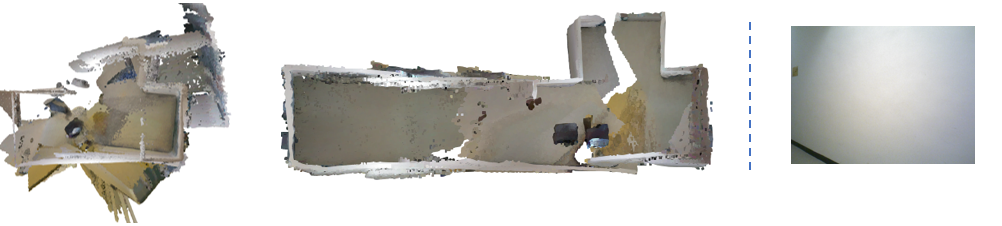}
     \put(20, -10) {\small (a)}
     \put(111, -10) {\small (b)}
    \put(205, -10) {\small (c)}
    \end{overpic}
    \vspace{-2mm}
\caption{Reconstruction results of a scene with a large expanse of featureless walls. (a) The reconstruction result by continuous camera tracking using ORB-SLAM visual odometry \cite{mur2017orb}; (b) our result by aligning partial scans; (c) a featureless wall that fails camera localization in this scene.}
\label{fig:featureless}
\vspace{-2mm}
\end{figure}

\para{Featureless scene reconstruction}For scenes that have a large expanse of featureless walls, it is very difficult for the existing methods to reconstruct them by continuous feature tracking. Fig. \ref{fig:featureless} shows the advantage of our method in reconstructing this kind of scene, while we directly align a set of partial scans caused by camera interruptions without using feature matching.

\begin{figure}[!htb]
\centering
    \begin{overpic}[width=\linewidth]{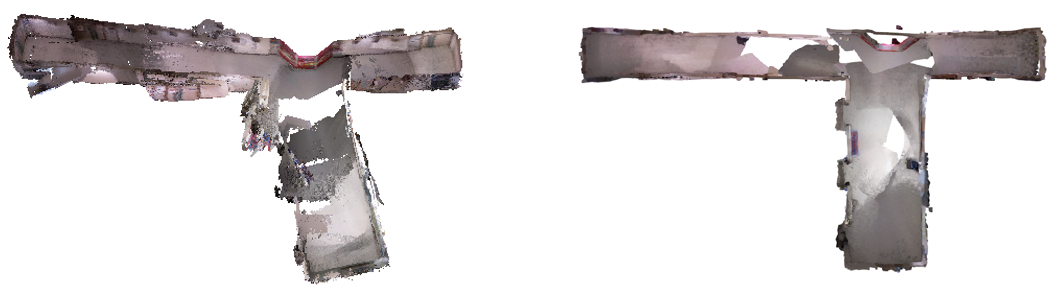}
    \end{overpic}
    \vspace{-6mm}
\caption{Reconstruction results of a large scene. Left: the result by continuous camera localization using ORB-SLAM visual odometry \cite{mur2017orb}; right: our result by aligning a set of partial scans. }
\label{fig:large_scene}
\vspace{-2mm}
\end{figure}

\para{Large scene reconstruction}As aforementioned, scanning a large scene region-by-region is easier than in a single pass due to the heavy workload, the accumulation error and the memory limit of a computer. Fig. \ref{fig:large_scene} shows the reconstruction results for a large scene using different strategies. In practice, we pay more efforts to maintain the uninterrupted scanning, but it still exhibits large accumulative errors. Instead, using region-based scanning, the scene is first divided into sub-regions and scanning each one separately is easier. Also, this strategy achieves better accuracy as illustrated.

\begin{figure}[!htb]
\centering
    \begin{overpic}[width=\linewidth]{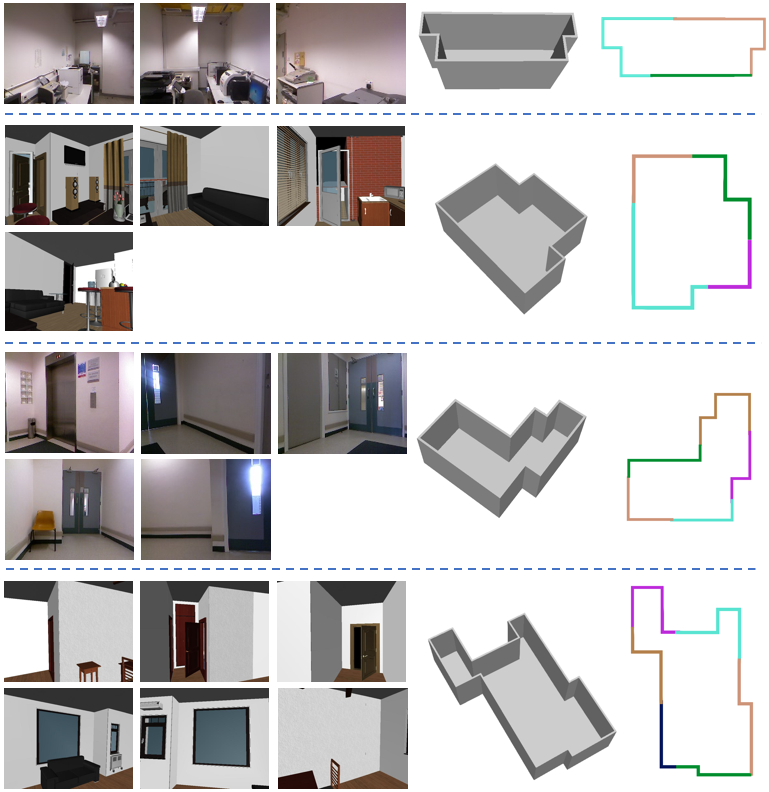}
    \end{overpic}
    \vspace{-5mm}
\caption{Room layout modeling and camera pose estimation by stitching sparse RGB-D frames. }
\label{fig:rgbd_results}
\vspace{-4mm}
\end{figure}

\para{Modeling from sparse input and down-sampling}The proposed method can recover a room layout from a small number of RGB-D images without adequate overlap, which can be used to model a scene given sparse input and down-sample the RGB-D stream in a scanning system (e.g., Matterport scanning system) for efficiency. As shown in Fig. \ref{fig:rgbd_results}, our method successfully aligns the RGB-D sequences and estimates the room layouts accordingly, which shows the ability of our method in modeling from sparse input.

\section{Conclusion}
In this work, we propose a novel approach to jointly align a set of partial reconstructions caused by camera interruptions and predict a room layout. Instead of relying on feature descriptor matching, our method is able to estimate the transformations of the partial 3D data without sufficient overlap, which is proved to be a challenge for the existing methods. Technically, we first estimate a local layout for each partial data and further formulate an optimal placement problem to combine these local layouts into a global loop closure under certain constraints. We have evaluated our algorithm quantitatively and qualitatively and compared it with the state-of-the-art methods, all of which demonstrate the effectiveness of our method on the alignment of small-overlapping partial scans as well as the global (partial) room layout estimation. 

\para{Acknowledgement}We thank the anonymous reviewers for their insightful comments. We are also grateful to Yasutaka Furukawa and Shiqing Xin for the inspiring discussions and valuable suggestions, and to Jiarui Wang for the data preparation. This work is supported by Hong Kong Innovation and Technology Support Programme (ITF ITSP) (ITS/457/17FP).

{\small
\bibliographystyle{ieee}
\bibliography{egbib}
}

\newpage







\section{Supplementary}
In this supplementary material, we include additional details for our Floorplan-Jigsaw paper. First, we show the detailed formulation of the boundary similarity analysis. Second, we describe the modifications to accommodate our method to RGB-D input. Furthermore, we show more quantitative results on our testing data.

\subsection{Boundary similarity analysis}
For a boundary point $\mathcal{P}$, given the closest point $\mathcal{P}'$ in the adjacent boundary of the next fragment, the probability of  $\mathcal{P}$ and  $\mathcal{P}'$ not belonging to a same object is computed by 
\begin{equation}
\setlength{\abovedisplayskip}{5pt} 
\setlength{\belowdisplayskip}{5pt}
P\left(\mathcal{P}, \mathcal{P}'\right)= \Phi\left(\frac{\Delta d}{\sigma}\right)-\Phi\left(\frac{-\Delta d}{\sigma}\right)\text{,}
\end{equation}
where $\Delta d$ is the distance between $\mathcal{P}$ and $\mathcal{P}'$. We consider a Gaussian measurement noise with standard deviation $\sigma$ and use its cumulative distribution function (CDF) denoted as $\Phi$ to compute the target area as the probability. Then the boundary similarity energy $E_b(f_l,f_k)$  when fragment $f_l$ is placed next to $f_k$ is defined as
\begin{equation}
\setlength{\abovedisplayskip}{5pt} 
\setlength{\belowdisplayskip}{5pt}
\frac{1}{M+N}\left(\sum_{\mathcal{P}_i\in B^r_{f_k}}\!\!P\left(\mathcal{P}_i,\mathcal{P}'_i\right)+\!\!\!\sum_{\mathcal{P}_{j}\in B^h_{f_l}}\!\!P\left(\mathcal{P}_j,\mathcal{P}'_j\right)\right)\text{,}
\end{equation}
where $M$ and $N$ are the number of the boundary points in $f_k$'s rear (denote as $B^r_{f_k}$) and $f_l$'s head (denote as $B^h_{f_l}$) respectively. This formulation gives a mismatch score  $E_b(f_k,f_l)$ ranging from 0 to 1, while it is set to 0.5 if either of the two fragments does not have enough boundary points ($M$ or $N$ $< 50$). We sum up the mismatch scores of all adjacent pairs to derive the final boundary similarity $E_b$.

\subsection{RGB-D images}
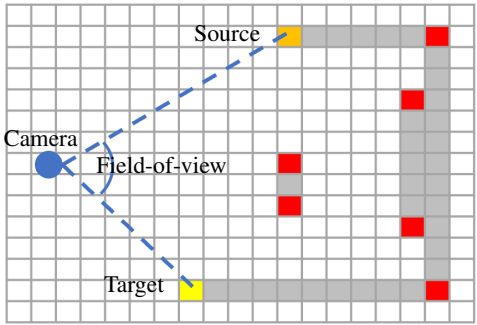
\begin{figure}[!htb]
\centering
\begin{overpic}[width=0.8\linewidth]{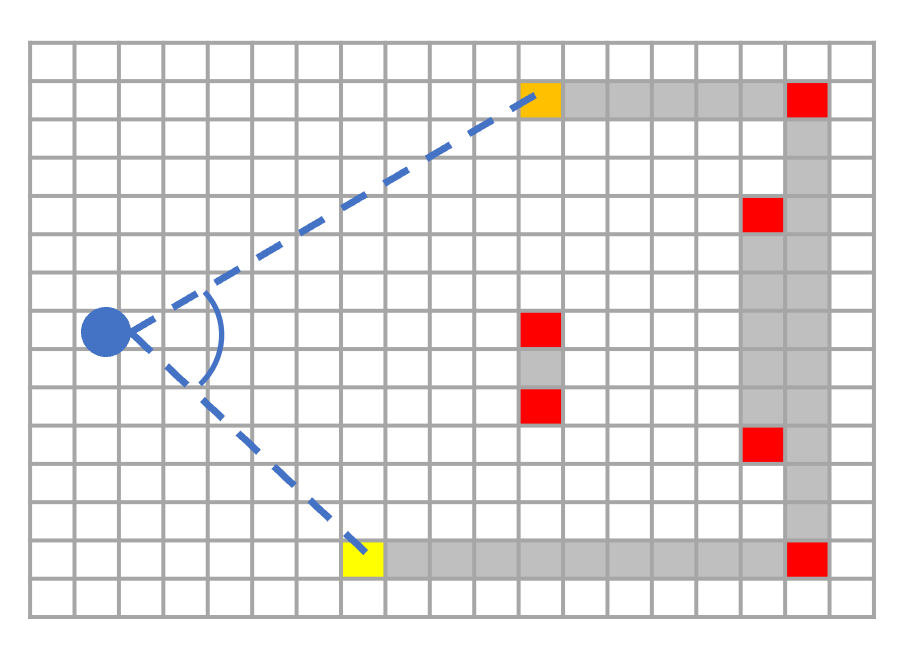}
        \put(5, 75) {\small Camera}
        \put(77, 115) {\small Source}
        \put(40,65) {\small Field-of-view}
        \put(43, 20) {\small Target}
\end{overpic}
\caption{We project the camera frustrum onto the floorplan plane and show the relationship between the source/target points and the field-of-view. }
\label{fig:fov}
\end{figure}
When the input are RGB-D frames, we modify the algorithm to make full use of the properties of image input in order to improve robustness.
First, the source and the target of the floorplan path can be efficiently determined by the image boundary, while we do not need to solve the \emph{ST-graph}. Fig.~\ref{fig:fov} shows the projection of the camera frustum onto the floorplan plane, where all the projected points in the scene will fall within the 2D FoV of the camera. The boundary of the scene (i.e., layout) should intersect with the 2D camera frustum. Therefore, the source and target points $p_i$ and $p_j$ of a floorplan path can be inferred by
\begin{equation}
    \arg\max\limits_{p_i,p_j\in\overline{P}} {\angle(p_i O_c p_j)},
\end{equation}
where $O_c$ is the camera position, $\overline{P}$ the set of candidate wall keypoints. 

To filter out the frames that have sufficient overlap, we detect ORB feature points for each image before running our algorithm. Any two images that have enough correct matches will be merged first and then the remaining frames can be considered as insufficiently overlapping. With these changes, the algorithm would be more robust to the input of RGB-D images. 

\subsection{Additional qualitative results}
The input to our method can be either partial point clouds or RGB-D images. Our testing data contains 101 scenes, among which 28 scenes are captured in the real-world and 73 scenes are synthesized from the SUNCG dataset. There are 22 scenes given as point clouds and 79 scenes given in the form of RGB-D images.

\begin{figure*}[!htb]
\centering
\vspace{-20mm}
\begin{overpic}[width=0.95\linewidth]{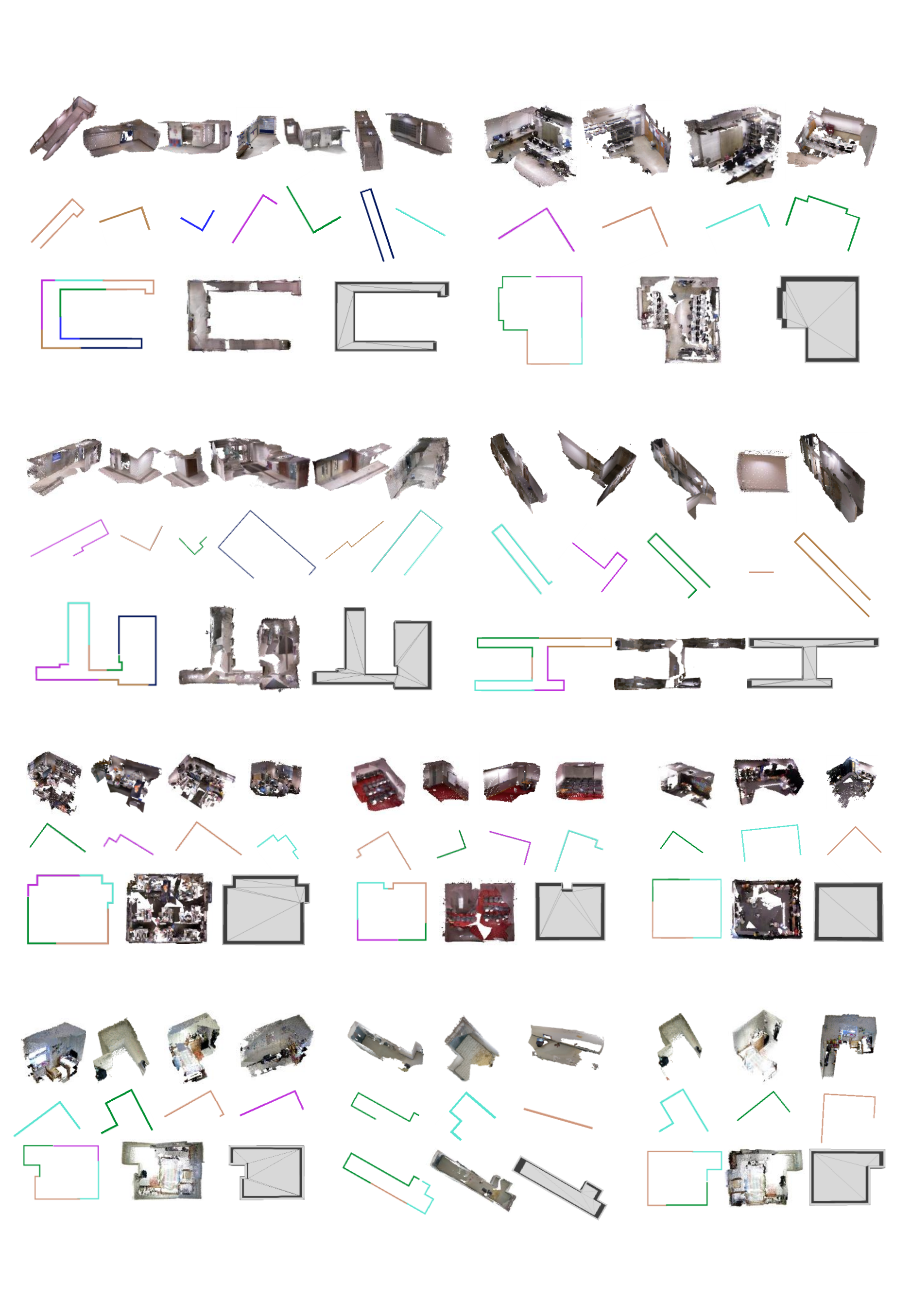}
\end{overpic}
\vspace{-15mm}
\caption{Additional qualitative results of partial scan alignment. We show each partial scan (first row of each sub-figure), the estimated local layout (second row of each sub-figure), the aligned global layout (first column of the last row of each sub-figure), the aligned point cloud (second column of the last row of each sub-figure) and the reconstructed layout model (third column of the last row of each sub-figure). }
\label{fig:appendix-pc}
\end{figure*}

\begin{figure*}[!htb]
\vspace{-20mm}
\centering
\begin{overpic}[width=\linewidth]{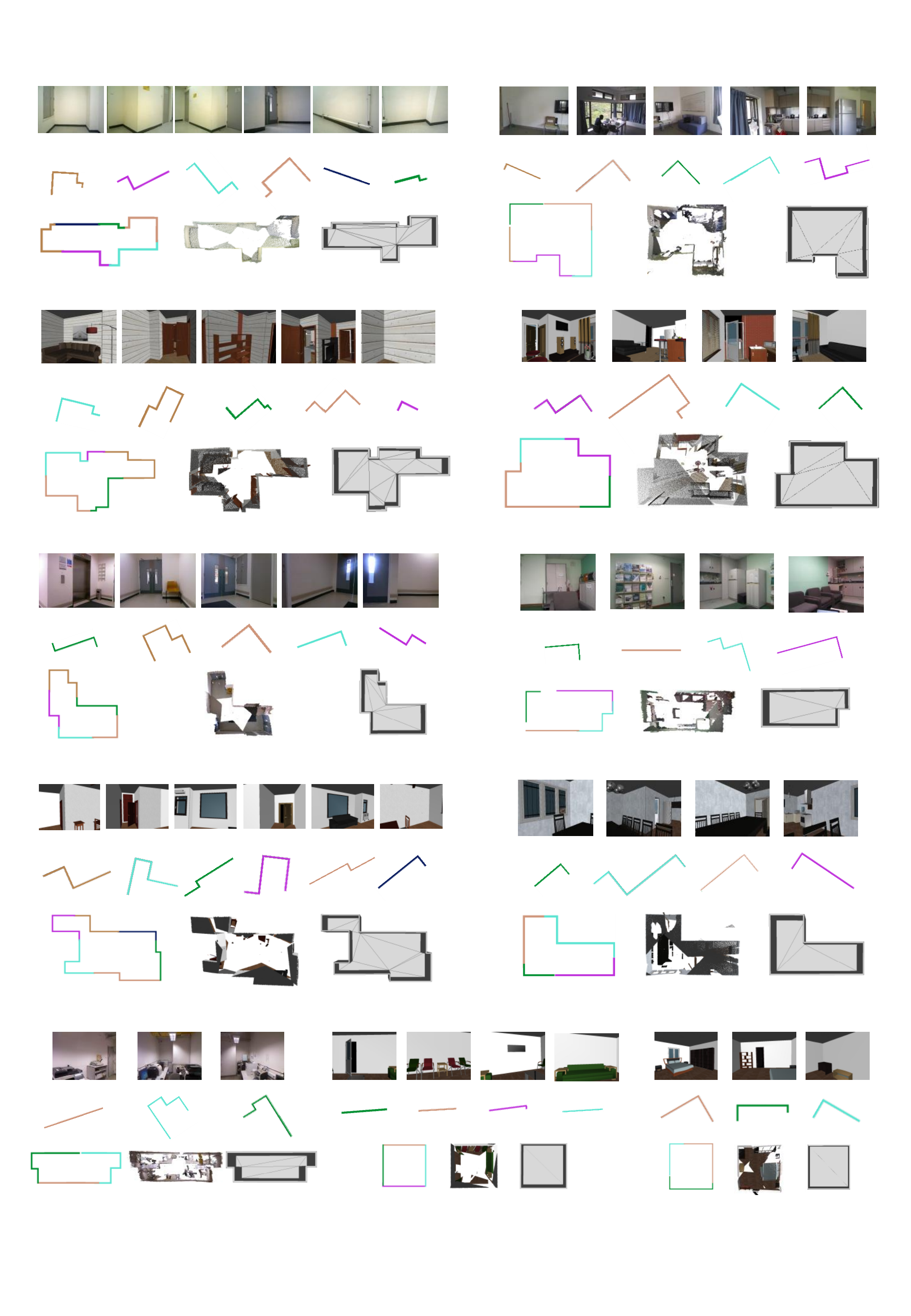}
\end{overpic}
\vspace{-25mm}
\caption{Additional qualitative results given RGB-D images as input. We show each RGB-D image (first row of each sub-figure), the estimated local layout (second row of each sub-figure), the aligned global layout (first column of the last row of each sub-figure), the aligned point cloud (second column of the last row of each sub-figure) and the reconstructed layout model (third column of the last row of each sub-figure). }
\label{fig:appendix-rgbd}
\end{figure*}

We show more detailed results for both point cloud input in Fig.~\ref{fig:appendix-pc} and RGB-D input in Fig.~\ref{fig:appendix-rgbd}. Our method is able to handle scenes with various sizes and layout complexities.

\end{document}